\documentclass[10pt,twocolumn,letterpaper]{article}

\usepackage{cvpr}
\usepackage{times}
\usepackage{epsfig}
\usepackage{graphicx}
\usepackage{amsmath}
\usepackage{amssymb}

\usepackage{bm}
\usepackage{subfigure}
\usepackage{amsthm}
\usepackage{multirow,array}
\usepackage{color}
\usepackage{algorithmic,algorithm}
\usepackage{lipsum}
\usepackage{booktabs}

\newcommand{\paraheading}[1]{\vspace{1em} \noindent \textbf{#1} \hspace{0.2em}}

\usepackage[breaklinks=true,bookmarks=false]{hyperref}

\cvprfinalcopy 

\def\cvprPaperID{****} 

\begin{document}

\title{CariGAN: Caricature Generation through Weakly Paired Adversarial Learning}

\author{Wenbin Li$^{\dag\ast}$,~Wei~Xiong$^{\ddag}$\thanks{Equal contribution},~Haofu~Liao$^\ddag$,~Jing~Huo$^\dag$,~Yang~Gao$^\dag$,~Jiebo~Luo$^\ddag$\\
$^\dag~$State Key Laboratory for Novel Software Technology, Nanjing University, China\\
$^\ddag~$Department of Computer Science, University of Rochester, USA\\
{\tt\small liwenbin.nju@gmail.com,~\{wei.xiong, haofu.liao\}@rochester.edu}\\
{\tt\small\{huojing, gaoy\}@nju.edu.cn, jluo@cs.rochester.edu}
}
\maketitle

\begin{abstract}
Caricature generation is an interesting yet challenging task. The primary goal is to generate a plausible caricature with reasonable exaggerations given a face image. Conventional caricature generation approaches mainly use low-level geometric transformations such as image warping to generate exaggerated images, which lack richness and diversity in terms of content and style. The recent progress in generative adversarial networks (GANs) makes it possible to learn an image-to-image transformation from data so as to generate diverse output images. However, directly applying GAN-based models to this task leads to unsatisfactory results due to the large variance in the caricature distribution. Moreover, some models require pixel-wisely paired training data which largely limits their usage scenarios. In this paper, we model caricature generation as a weakly paired image-to-image translation task, and propose CariGAN to address these issues. Specifically, to enforce reasonable exaggeration and facial deformation, facial landmarks are adopted as an additional condition to constrain the generated image. Furthermore, an image fusion mechanism is designed to encourage our model to focus on the key facial parts so that more vivid details in these regions can be generated. Finally, a diversity loss is proposed to encourage the model to produce diverse results to help alleviate the ``mode collapse'' problem of the conventional GAN-based models. Extensive experiments on a large-scale ``WebCaricature'' dataset show that the proposed CariGAN can generate more plausible caricatures with larger diversity compared with the state-of-the-art models.
\end{abstract}

\section{Introduction}
\begin{figure}[tpb]
      \centering
      \includegraphics[width = 0.47\textwidth]{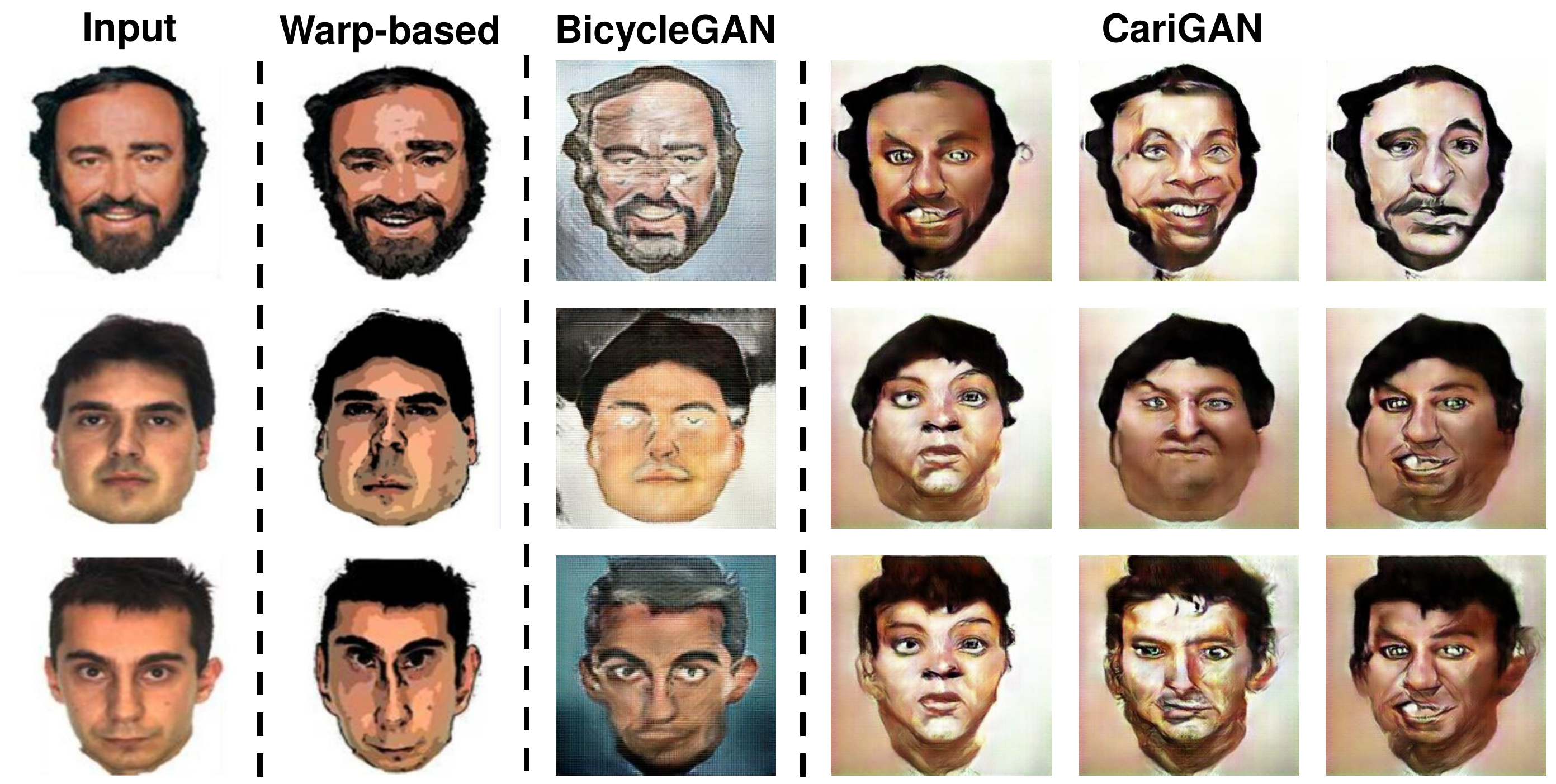}
      \caption{Results by different types of models for caricature generation. From left to right: input face images (Col. $1$), geometric deformation based model (Col. $2$), BicycleGAN (Col. $3$) and our CariGAN model (Col. $4\!-\!6$), which are produced by adapting different poses to the same input image.}
      \label{fig-priorwork}
\end{figure}

Caricature is an artistic creation produced by exaggerating some prominent characteristics of a face image while preserving its identity. Caricatures are widely used in social media and daily life for a variety of purposes. For example, it can be used as the profile image or to express certain emotions and sentiments on social networks. Due to the prosperity of social media, automatic caricature creation becomes an increasingly attractive research problem. In this paper, given an arbitrary face image of a person, our primary goal is to generate satisfactory or plausible caricatures of that person with reasonable exaggerations and an appropriate caricature style.

To this end, we identify and define four key aspects  that need to be taken into account for caricature generation: 
\begin{itemize}
\vspace{-2mm}
\item \textbf{Identity Preservation:} The generated caricature should share the same identity as the input face;
\vspace{-2mm}
\item \textbf{Plausibility:} The generated caricature should be visually satisfactory or plausible; the style of the generated image should be consistent with normal cartoons or caricatures;
\vspace{-2mm}
\item \textbf{Exaggeration:} Different parts of the input face should be deformed in a reasonable way to exaggerate the prominent characteristics of the face;
\vspace{-2mm}
\item \textbf{Diversity:} Given an input face, diverse caricatures with different styles should be generated.
\vspace{-2mm}
\end{itemize}

Several previous studies~\cite{akleman1997making,iwashita1999expressive,koshimizu1999kansei,brennan2007caricature,LiuCXGG09,sadimon2010computer} have made attempts to solve this problem. These studies mainly focus on the generation of sketch caricature~\cite{ChenXSZZ01,liang2002example,mo2004improved}, black-white illustration caricature~\cite{gooch2004human}, and outline caricature~\cite{fujiwara2000web}. Most of them adopt low-level image transformations and computer graphics techniques~\cite{akleman2000making,liu2006mapping,tseng2012colored,yang2016example} to generate new images. They are either semi-automatic or complicated with multiple stages, making it difficult to be applied to large-scale caricature generation applications. Moreover, although they can generate correct deformations on some facial parts, their results are usually visually unappealing, \emph{e.g.}, lacking of rich colors and vivid details. As shown in the second column of Figure~\ref{fig-priorwork}, the conventional low-level geometric deformation based approaches~\cite{tseng2012colored} can only generate one specifically exaggerated caricature for one input face. Often, the content, texture and style of the generated caricature are plain and less interesting.

Recently, with the progress of conditional generative adversarial networks (GANs)~\cite{goodfellow2014generative,mirza2014conditional,radford2015unsupervised} and their success in image generation, image translation and editing tasks~\cite{chen2016infogan, springenberg2015unsupervised, isola2016image,CycleGAN2017,zhu2017toward, berthelot2017began}, it is possible to use a GAN model to learn transformations from data itself to produce plausible caricatures from the input face images. However, although typical GAN-based models such as Pix2Pix~\cite{isola2016image} can generate realistic images, directly applying these models to this task fails to produce satisfactory outputs. Most of the previous methods cannot address all of the four key aspects together. Quite often, the generated image is almost visually the same as the input face with only minor changes in color, lacking sufficient exaggerations in facial parts. As shown in the third column of Figure~\ref{fig-priorwork}, there is almost no exaggeration of facial features, which does not satisfy the primary goal of caricature generation. In addition, many GAN-based image-to-image translation models require strictly paired training images, \emph{i.e.,} the transformation should be a bijective pixel-to-pixel mapping. However, these paired data are quite difficult to obtain. For caricature generation, using such pixel-wisely paired data is not feasible for practical purposes. 

Inspired by the power of the conditional GANs, this paper proposes an end-to-end model named CariGAN to solve the problems encountered by the conventional GAN models. The goal is to address as much as possible the four key aspects of caricature generation.

Due to the difficulty in obtaining strictly paired training data, we introduce a new setting for training GANs, \emph{i.e.}, \emph{\textbf{weakly paired training}}. Specifically, one pair of input face and the ground-truth caricature only share the same identity but has no pixel-to-pixel or pose-to-pose correspondence. This setting is much more challenging than the pixel-wisely paired training setting. We will describe this setting in detail in Section~\ref{weak_pair_section}.

Furthermore, as shown in Figure~\ref{fig-priorwork}, although conventional GAN-based models such as BicycleGAN~\cite{zhu2017toward} can produce caricatures with correct identities, they fail to produce reasonable exaggerations. It is worth emphasizing that the exaggeration is a vital aspect to make a vivid caricature. In our model, we retain the advantage of conventional models, \emph{i.e.,} employing a U-net as the generator to keep the identity unchanged during the transformation. In addition, we introduce a facial mask as a condition of GAN to precisely guide the deformations of faces, so that the generated images can have reasonable exaggerations.

For the plausibility issue, although the GAN-based models can produce plausible images by forcing the distribution of the generated caricatures to be close to that of the ground truth, there are still many artifacts that decrease the degree of plausibility. To enhance the plausibility of the generated caricatures, a new \emph{\textbf{image fusion mechanism}} is proposed. By adopting this mechanism, we can encourage the model to concentrate on refining not only the global appearance, but also the important local facial parts of the generated caricature images.

Finally, many conditional GAN models suffer from the so-called ``mode collapse'' problem, \emph{i.e.}, different inputs, especially random noise, can be mapped to the same mode \cite{isola2016image}. The diversity of the outputs will be greatly reduced due to this problem. To address this problem, a novel \emph{\textbf{diversity loss}} is proposed to enforce that the input random noise should play a more important role in generating the styles and colors of the generated caricatures.

In summary, the main contributions of this paper are as follows:
\begin{itemize}
    \vspace{-2mm}
    \item We introduce \emph{a new weakly paired training setting for GANs} and propose a CariGAN model that can successfully generate plausible caricatures under this challenging setting. 
    \vspace{-2mm}
    \item We propose \emph{a new image fusion mechanism} to encourage the model to focus on both the global and local appearance of the generated caricatures, and pay more attention to the key facial parts. 
    \vspace{-2mm}
    \item We propose \emph{a novel diversity loss}  to encourage our model to generate caricatures with larger diversity in style, color and content.
    \vspace{-2mm}
\end{itemize}
The rest of this paper is organized as follows: Section~\ref{related_work} introduces the related work on caricature generation, conditional generative adversarial networks and multimodality encoding in GANs. Section~\ref{our_model} introduces the proposed model in details. Experimental settings and results of different models are provided in Section~\ref{experiments}, and the last section concludes the whole paper.

\section{Related work}
\label{related_work}
\subsection{Caricature Generation}
Early work on caricature generation mainly focuses on low-level image processing and computer graphics approaches. Typical process for this kind of approaches can be summarized as follows: (1) detect facial feature points (\emph{i.e.,} facial landmarks) and extract facial sketch from an input face; (2) find the distinctive characteristic and exaggerate the facial shape; (3) warp the original face image to the exaggerated one to get a caricature.

There are two major types of these earlier work: rule-based methods and example-based methods. Rule-based methods generate caricatures by simulating the rules of caricature drawing, \emph{i.e.,} the notion of ``exaggeration the difference from the mean'' (EDFM). In general, an average face or a standard face model is taken as a reference, and then the difference is exaggerated. Representative methods include~\cite{brennan2007caricature,mo2004improved,gooch2004human,liao2004automatic}. In \cite{liao2004automatic}, Chiang \emph{et al.} formalized the caricature generation into a metamorphosis process to generate caricatures by leveraging one caricature as a reference. In~\cite{mo2004improved}, Mo \emph{et al.} extended the notion of EDFM by considering both feature DFM (Difference From Mean) and feature variance. Unlike rule-based methods, example-based methods rely on a face-caricature dataset and generate caricatures based on similar examples. For example, Liang \emph{et al.}~\cite{liang2002example} proposed a prototype-based exaggeration model by analyzing the correlation between face-caricature pairs. Liu \emph{et al.}~\cite{liu2006mapping} adopted principal components analysis (PCA) to get the principal components, and then employed support vector regression (SVR) to learn a mapping model to generate caricatures. Recently, Yang \emph{et al.}~\cite{yang2016example} took both the spatial relationship among facial components and the shape of facial components, into account and proposed a new example-based method. Zhang \emph{et al.}~\cite{ZhangDMMLHHD17} proposed a data-driven framework for generating cartoon faces by selecting and assembling facial components from a database.
\begin{figure*}[!thp]
      \centering
      \includegraphics[width = 0.85\textwidth]{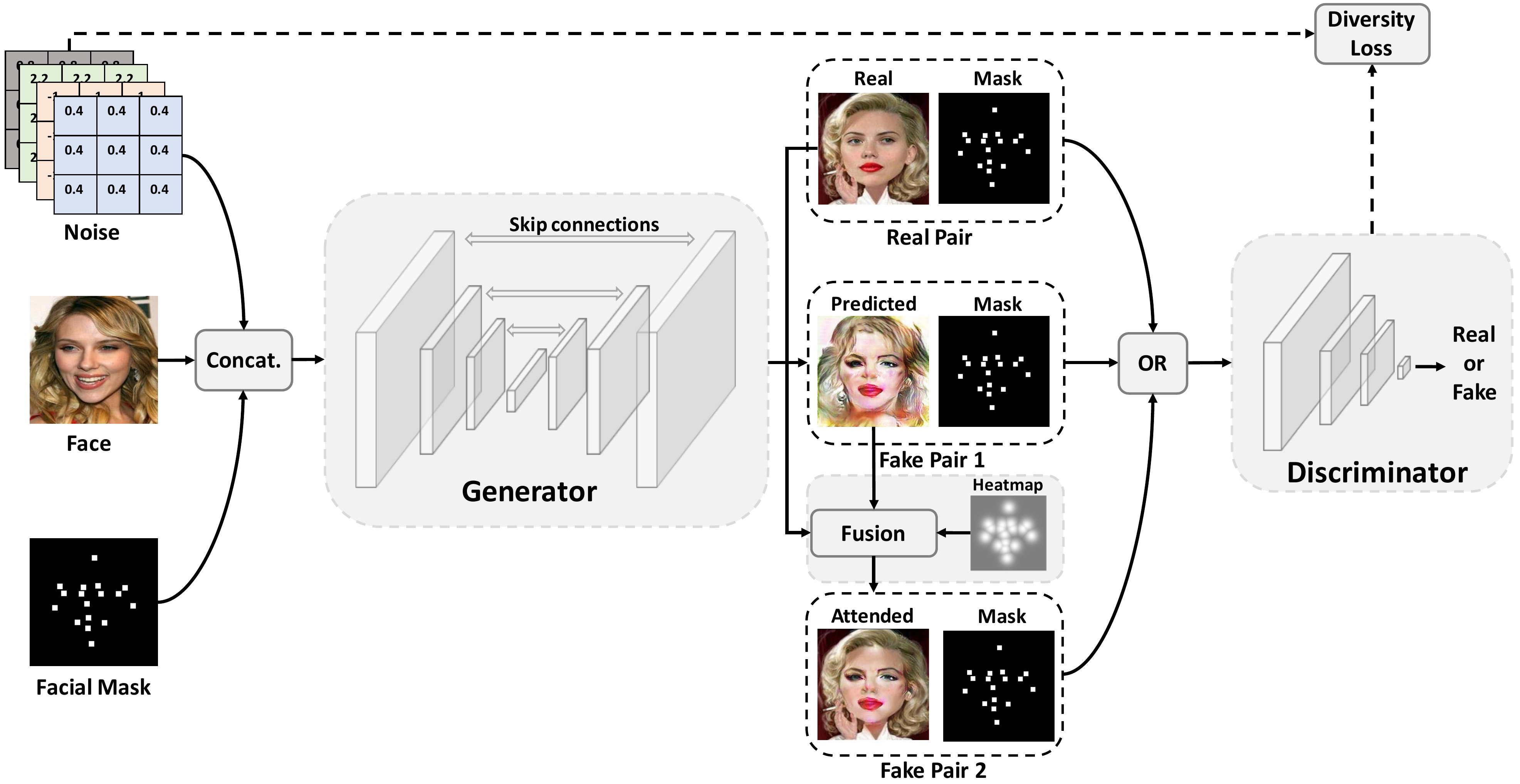}
      \caption{The overall framework of the proposed CariGAN. The input of the generator is a concatenation of a random noise map, a face image and a facial mask. The input data is fed into a U-net generator to generate a fake caricature. Then, using our \emph{image fusion mechanism}, we combine the ground-truth and the generated fake caricature to get an additional fused fake caricature. These caricatures are concatenated with the input facial mask, respectively, and fed into the discriminator. The discriminator then tries to distinguish the real from the fakes. In addition to the adversarial loss, a \emph{diversity loss} is proposed to constrain the outputs to be more diverse in style and content.}
      \label{fig-framework}
\end{figure*}

\subsection{Conditional Generative Adversarial Networks}
Caricature generation can be seen as an image translation problem and thus can be modeled with conditional generative adversarial networks (cGANs)~\cite{goodfellow2014generative, mirza2014conditional}. A conditional GAN takes a random noise and some prior knowledge as inputs to generate data whose conditional distribution is similar to the one of the ground-truth data. Recently, cGANs~\cite{arjovsky2017wasserstein, mathieu2015deep,xiong2017learning,chen2017show,reed2016generative,mescheder2018training,ma2018gan} have shown great capacity in learning transformations from data and generating realistic images. Typical supervised models such as Pix2Pix~\cite{isola2016image} and BicycleGAN~\cite{zhu2017toward} perform well on the image-to-image translation problem, especially when the input image and the output image have a pixel-wise correspondence. To relieve the requirement of strictly paired training data, CycleGAN~\cite{CycleGAN2017}, DiscoGAN~\cite{kim2017learning}, and Dual GAN~\cite{yi2017dualgan} demonstrated that such tasks can even be accomplished in an unsupervised way.

However, directly applying these supervised or unsupervised GAN models to the caricature generation task may fail to generate plausible caricatures due to the weakly paired nature of our task, \emph{e.g.,} different facial poses between face images and caricatures, and varying degrees of exaggeration and deformation among facial components in caricatures. To tackle this problem, our CariGAN model is built not only on condition of the input face image, but also a facial mask which indicates the landmarks of the target caricature. Through the condition of facial mask, the generated caricature can be encouraged to have a similar exaggeration and viewpoint as the ground-truth caricature. Similar to our model, some GAN-based models use an additional person pose mask to guide the generation process. For example, Ma \emph{et al.}~\cite{ma2017pose} used a person pose to guide a two-stage GAN-based model to generate realistic person images. In the first stage, it adopted a reconstruction loss to generate a coarse image which was then refined in the second stage by a GAN model. One major difference between their model and ours lies in that reconstruction plays a key role in their model, which may lead to blurry results~\cite{isola2016image}. On the contrary, our model takes full advantage of adversarial learning and is able to generate more sharp images. Another difference is that they use multiple stages which is more complicated, while our model is an end-to-end one stage model.

Another closely related work is~\cite{zheng2017photo}, which is also based on GANs for caricature generation. The major differences are that: (1) our model is trained on weakly paired face-caricature images, while~\cite{zheng2017photo} requires strictly paired images with the same facial viewpoint for training; (2) our model is conditioned on a face image and a facial mask, which can control the exaggeration of the output, while \cite{zheng2017photo} is only conditioned on the input face, lacking the ability to control the exaggeration.

\subsection{Mutlimodality Encoding in GANs}
One major issue regarding cGANs is the ``mode collapse'' problem~\cite{goodfellow2016nips}. In order to relieve this problem, the key point is on how to learn richer modalities of the outputs and avoid multiple inputs being mapped to the same output. Some prior studies addressing this problem \cite{zhu2017toward,larsen2015autoencoding,donahue2016adversarial,belghazi2018mutual, miyato2018spectral, huang2018munit, lee2018stochastic} have been proposed. One simple and effective way to alleviate this problem is to use a latent code as an additional input to explicitly encode the modes. For example, a one-hot vector representing the facial viewpoint is introduced as an input to generate faces with different poses \cite{tran2017disentangled}. In our work, we use a facial mask to guide the generation of caricatures.

Another typically applied approach to relieve the ``mode collapse'' problem is to enforce a tight connection between the latent codes and the output data. A few previous studies have investigated this idea by introducing an additional encoder to map the generated image back to the input random noise, so that the mapping from the random noise to the output can be bijective \cite{zhu2017toward,larsen2015autoencoding,donahue2016adversarial}. However, the encoder brings additional computation, and the simultaneous optimization of the generator and encoder is non-trivial. We provide a new perspective to solve this problem. In addition to using the facial mask as a guidance, we enforce the differences between the output images to be a linear function of the differences between the input random noises, so that the change of noise can greatly influence the styles of the output images.

\section{Our Model}
\label{our_model}
As illustrated in Figure~\ref{fig-framework}, CariGAN takes a face image, a facial mask and a random noise as inputs. It then tries to generate a plausible caricature that has the same identity with the input face and meaningful exaggeration as indicated by the input facial mask. To produce satisfactory caricatures, CariGAN uses a generative adversarial network to model the translation from a face to a caricature. To encourage the model to generate realistic caricatures with more reasonable exaggerations, we introduce an image fusion mechanism to this model to focus more on the important facial parts of the generated image. We also design a diversity loss to address the ``mode collapse'' problem. The diversity loss enforces the differences between the output images to be a linear function of the differences between the input random noises. 

\subsection{Adversarial Learning with Weak Pairs}
\label{weak_pair_section}
\paraheading{Weakly paired training setting} Let $(x, y)$ be a pair of training data, where $x$ represents the input face image, and $y$ represents the corresponding ground-truth caricature. $x$ and $y$ are of $256\!\times\!256$ resolutions and belong to the same person. It should be noted that they are not pixel-wisely or pose-wisely paired. This setting is quite different from the conventional paired training setting, where the input image and the ground-truth are usually pixel-wisely and bijectively mapped~\cite{isola2016image,zhu2017toward}. This is because that there are multiple face images and various caricatures with different artistic styles for one person, which means that one face image can be paired with multiple caricatures and one caricature can also correspond to multiple face images of the same identity. Thus, in an input pair, the face image and caricature can have totally different viewpoints (\emph{i.e.,} facial poses), which makes the task extremely challenging. In addition to the viewpoint, there is no pixel-wise correspondence between the faces and the caricatures inherently, as many facial parts are exaggerated. Hence, we call this pair a \emph{weak pair}, and define this setting as a new training setting, namely \emph{weakly paired training}, in the image-to-image translation task. 

\paraheading{Adversarial loss} The goal of our task is to map an input face image $x$ to a caricature image $\hat{x}$ such that the distribution of variable $\hat{x}$ is close to that of the weakly paired ground-truth caricature image $y$. To this end, we build a CariGAN model based on the cGANs to handle this image-to-image translation task. Our CariGAN is composed of a generator $G$ and a discriminator $D$. With an input face image $x$ and a random noise $z$, $G$ tries to generate a caricature image $\hat{x}$. The goal of $G$ is to make $\hat{x}$ as plausible as possible, so as to fool the discriminator $D$, while the discriminator tries to distinguish the generated image $\hat{x}$ and the ground-truth $y$. Specifically, following the usage of noise variable in BicycleGAN~\cite{zhu2017toward}, we first sample a noise vector of length $4$ from a Gaussian distribution. Then it is duplicated $256\times 256$ times in the spatial locations to get a $4\times 256 \times 256$ noise map $z$. We then directly concatenate $x$ and $z$ as the input of our generator. The adversarial loss of such a conditional GAN can be formulated as:
\begin{equation} 
	\mathcal{L}_{adv}={\mathbb{E}}\big[\log D\left(y\right)\big]+{\mathbb{E}}\big[\log\left(1-D\left(G\left(x,z\right)\right)\right)\big] \,.
\end{equation}

\paraheading{Facial mask as an additional condition} Unfortunately, only conditioning on the input face makes it difficult to learn reasonable exaggerations in the output caricature for the following reasons: (1) One input face can actually be mapped to caricatures with arbitrary exaggerations. This uncertainty may confuse the generator. (2) Although the input noise can be used to model a wider distribution, it is difficult to encode viewpoints, exaggerations and styles at the same time.

To reduce the uncertainty, we use a facial mask $p$ as an additional condition and feed it into the generator $G$ along with $x$. We encourage the model to generate a caricature that has similar exaggerations as indicated by this mask. The facial mask is a binary image composed of $17$ facial landmarks. In the mask, each landmark is represented by a $11\times 11$ square block and we fill the pixels in the blocks with ones and the background pixels with zeros. The facial mask can encode two aspects of a face. The first aspect is the exaggeration on local facial parts, such as eyes, mouth \emph{etc}. The second one is the viewpoint of the whole face.

During training, we directly use the facial mask of the ground-truth $y$ as input and constrain the output $\hat{x}$ of the generator to be similar to $y$ with regard to facial exaggeration and viewpoint. In this way, the major appearance of the output image is roughly determined, except for some variations on the styles, textures and colors. The success of the previous conditional GAN models~\cite{mirza2014conditional,isola2016image,CycleGAN2017} has indicated that the random noise sampled from a Gaussian distribution is able to model the variation of different styles. Hence, we also use random noise to encode the style of the generated caricature. In fact, we use the facial mask as an additional condition for both the generator and the discriminator. Specifically, we directly concatenate $x$, $p$ and $z$ to form an $8$-channel map as the input of our generator. The input of the discriminator is a concatenation of $(\hat{x}, p)$ or $(y, p)$. With facial mask as an additional input, the adversarial loss of our model is as follows:
\begin{equation} 
	\mathcal{L}_{adv}\!=\!{\mathbb{E}}\big[\log D\left(p, y\right)\big]\!+\!{\mathbb{E}}\big[\log\left(1\!-\!D\left(p, G\left(x, p, z\right)\right)\right)\big] \,.
\end{equation}

As the distribution of the generated fake pair $(\hat{x}, p)$ is encouraged to be close to the distribution of the real pair $(y, p)$, the generated image $\hat{x}$ is not only enforced to have similar appearance of the ground-truth $y$, but also enforced to have a similar exaggeration as indicated by $p$. If we only use $p$ as a condition in $G$ and ignore it in the discriminator $D$, then the $\hat{x}$ is only constrained to mimic the distribution of $y$. The input pose condition tends to be ignored during training in this case.

\paraheading{Content loss} Previous work on image-to-image translation~\cite{isola2016image} shows that combining the pixel-wise $\ell_1$ loss between the generated fake image and the ground-truth can boost the performance of cGANs. Although in our task, pixels of the ground-truth $y$ and the generated image $G(x, p, z)$ are not an bijective mapping, we discover that using an $\ell_1$ loss can stabilize the training of GANs. Hence, we also use this pixel-wise loss as a constraint for the content of the generated image. The content loss is formulated as:
\begin{equation}
	\label{l1_loss}
	\mathcal{L}_{con} =  \left\|y - G\left(x, p, z\right)\right\|_1 \, ,
\end{equation}
where $\|\cdot\|_1$ denotes $\ell_1$ norm of matrix.
\begin{figure}[!tbp]
      \centering
      \includegraphics[width = 0.46\textwidth]{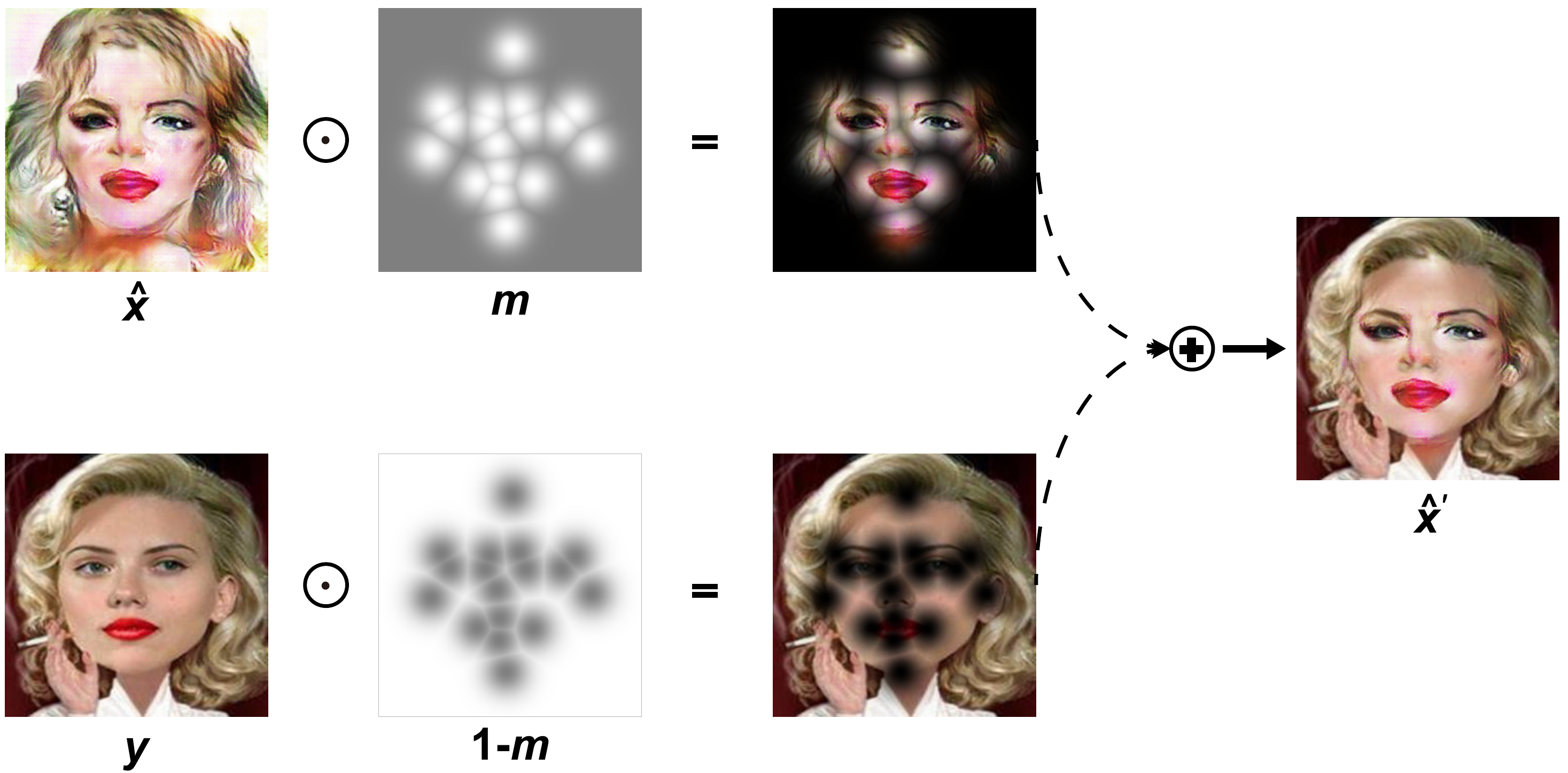}
      \caption{Illustration of the image fusion procedure. }
      \label{fig-attention}
\end{figure}

\subsection{Focus on Important Local Regions}
Although the conditional GAN is able to generate visually appealing images, there are still many local artifacts in the output images such as the absence of eyes. The reason may be that the conventional conditional GANs only constrain that the global appearance of the generated image should look like real caricatures on average, but it cannot guarantee that each local facial part is present and realistic. To encourage the model to generate reasonable facial parts, we propose a new image fusion (IF for short) mechanism to force the model to focus more on important local regions. We fuse the background parts of the ground-truth and the generated key local parts of the generated fake images to create new additional fake images. The basic idea is illustrated in Figure~\ref{fig-attention}.

Specifically, we use the input facial mask $p$ as a guidance for selecting the important regions. We create a Gaussian blob for each landmark in the facial mask and obtain a one-channel heatmap $m$. Using this heatmap, we replace the regions of the ground-truth $y$ around the landmarks with the regions of the generated image $\hat{x}$, and keep the other unimportant parts such as the background pixels unchanged. In this way, we generate an additional fused fake image $\hat{x}'$, which is formulated as:
\begin{equation} 
	\hat{x}' =  m \odot \hat{x} + (1-m) \odot y \, ,
    \label{attention_eq}
\end{equation}
where $\odot$ denotes the pixel-wise multiplication. With $\hat{x}'$ generated, it is fed into the discriminator $D$, which tries to distinguish not only $\hat{x}$ from $y$ but also $\hat{x}'$ from $y$. The adversarial loss is now changed to:
\begin{equation} \small
\begin{split}
	\mathcal{L}^{IF}_{adv}={\mathbb{E}}\big[\log D\left(p, y\right)\big]&+\frac{1}{2} {\mathbb{E}}\big[\log\left(1-D\left(p, \hat{x}\right)\right)\big]\\
& + \frac{1}{2}{\mathbb{E}}\big[\log\left(1-D\left(p, \hat{x}'\right)\right)\big] \, ,
\end{split}
\end{equation}
where $\hat{x}$ is the generated fake caricature by generator $G$, \emph{i.e.,} $\hat{x}=G(x,p,z)$, and $\hat{x}'$ is the additional fake caricature constructed by our image fusion module according to Eq.~(\ref{attention_eq}). Specifically, $\hat{x}$ and $\hat{x}'$ have the same weight, \emph{i.e.,} $0.5$, and both of them try to fool the discriminator $D$, while $D$ tries to distinguish them from the ground-truth $y$. 

The image fusion mechanism can improve not only the quality of the generated caricature in global appearance, but also the quality of its local appearance. On one hand, the discriminator distinguishes $\hat{x}$ from $y$, forcing the generator to produce images that mimic the global appearance of the ground-truth. On the other hand, since most of the parts of $\hat{x}'$ is exactly the same as $y$, the discriminator needs only to judge whether the focused regions look realistic. It then encourages the generator to pay more attention to the local facial parts and try to improve them to fool discriminator.

With the image fusion mechanism introduced, the content loss is modified accordingly. Our model is encouraged to focus more on the important regions, so the content loss is modified to the following form:
\begin{equation} 
	\mathcal{L}^{IF}_{con} =  \left\|(y - G\left(x, p, z\right)) \odot m \right\|_1 \, ,
\end{equation}
where $m$ is the heatmap created from the facial mask $p$. Compared with Eq.~(\ref{l1_loss}), this heatmap guided content loss can encourage the network to put more efforts on generating the important facial parts, such as the eyes, mouth, nose and so on. However, in the experiments we discover that the two losses have no significant difference on influencing the performance of the generated caricatures, but this loss does make the model slightly more stable than Eq.~(\ref{l1_loss}) during the training stage. 
\begin{algorithm}[!tp]
	\caption{The training procedure of the CariGAN model.}
	\label{algorithm1}
	\begin{algorithmic}
    \STATE Set learning rates $\rho_d$ and $\rho_g$ for the generator $G$ and discriminator $D$, respectively.
    \STATE Initialize the parameters $\theta_d$ of $D$ and $\theta_g$ of $G$.
		\FOR {number of iterations}
        \STATE \textbf{Updating $\theta_d$ while fixing $\theta_g$:}
		\STATE \quad Sample a batch of weakly paired training data {\small$\{(x^{(1)}, y^{(1)}), (x^{(2)}, y^{(2)}), \ldots, (x^{(N)}, y^{(N)})\}$}.
		\STATE \quad Get the pose {\small$\{ p^{(1)}, p^{(2)}, \ldots, p^{(N)} \}$} of the ground-truth.
        \STATE \quad Generate fake samples {\small$\{\hat{x}^{(1)},\hat{x}^{(2)}, \ldots, \hat{x}^{(N)}\}$} from {\small$G$}.
        \STATE \quad Generate fused fake samples {\small$\{\hat{x}'^{(1)}, \hat{x}'^{(2)}, \ldots, \hat{x}'^{(N)}\}$} from {\small$G$}.
		\STATE \quad {\small$\theta_d := \theta_d + \rho_d \nabla_{\theta_d}\dfrac{1}{N}\sum_{n=1}^N \mathcal{L}^{IF}_{adv}(\hat{x}'^{(n)}, y^{(n)}, p^{(n)})\,.$}
        \STATE \textbf{Updating $\theta_g$ while fixing $\theta_d$:}
		\STATE \quad {\small$\theta_g := \theta_g -\rho_g \nabla_{\theta_g}\dfrac{1}{N}\sum_{n=1}^N \mathcal{L}(\hat{x}'^{(n)}, y^{(n)}, p^{(n)})\,.$}
		\ENDFOR
	\end{algorithmic}
\end{algorithm}

\subsection{Diversity Loss}
In our proposed model, the random noise controls the colors and styles of the images. However, in practice, the proposed model may suffer from the ``mode collapse'' problem, \emph{i.e.}, the input noise may not able to affect the final results. 

To address the ``mode collapse'' problem, we propose a diversity loss to force our model to generate images with larger diversity. The basic idea of the diversity loss is to encourage the difference between two fake caricatures generated from two different noises (but with the same input face and facial mask) to be a linear function of the difference between these two noises. Suppose the generator is given a human face image $x$ and a binary pose mask $p$, but with two different noise $z_1$ and $z_2$. The generator outputs two fake caricatures, \emph{i.e.,} $\hat{x_1}$ and $\hat{x_2}$ for these two inputs, respectively. We have: $\hat{x_1} = G(x, p, z_1)$, $\hat{x_2} = G(x, p, z_2)$.

We then extract features of these two fake caricatures from the last convolutional layer of the discriminator $D$. Denote the extracted feature as $f_1 = D(\hat{x_1}, p)$, $f_2 = D(\hat{x_2}, p)$. The extracted feature encodes the identity, pose as well as style of the generated image. However, as the two features are extracted from two fake caricatures with the same identity and viewpoint, it is reasonable to treat the difference between these two features as the difference between the styles and other unimportant attributes. We therefore force the difference between the two features to be a linear function of the difference between the two input noises. In this way, the diversity of styles can be explicitly controlled by the input noise. Our diversity loss is formulated as:
\begin{equation} 
	\mathcal{L}_{div} =  \left(\frac{\|f_1 - f_2\|^2_2}{\|f_1\|^2_2 + \|f_2\|^2_2} - \frac{\|z_1 - z_2\|^2_2}{\|z_1\|^2_2 + \|z_2\|^2_2}   \right)^2 \,,
\end{equation}
where the difference of features and noises are normalized by the feature norms and noise norms respectively to have a similar magnitude. The overall loss of our proposed CariGAN model can be formulated as:
\begin{equation}
	\mathcal{L} = \mathcal{L}^{IF}_{adv} + \mathcal{L}^{IF}_{con} +  \mathcal{L}_{div} \, .
\end{equation}
To make our approach more understandable, we summarize the whole training procedure of CariGAN in Algorithm \ref{algorithm1}.

\section{Experiments}
\label{experiments}

\subsection{Basic Settings}
\paraheading{Dataset} All the experiments in this study are performed on the WebCaricature dataset~\cite{huo2017webcaricature}. The WebCaricature dataset contains $5974$ photograph and $6042$ caricature images of $252$ celebrities, which is currently the largest caricature dataset. Images of $200$ celebrities are used for training and the rest $52$ celebrities are hold out for testing. All the images are aligned according to the provided facial landmarks as follows: (1) rotate each image to make two eyes in a horizontal line; (2) resize each image to guarantee the distance between two eyes of 75 pixels; (3) crop the primary facial part as the face image and resize it to $256\times256$. Moreover, random flip is performed for augmentation. We construct weak pairs completely on the training set for training, obtaining $562,965$ weakly paired face-caricature images. During training, we randomly select a pair of face and caricature images as the input and ground truth, respectively. Each face or caricature image is associated with $17$ manually annotated facial landmarks from which we generate a binary mask $p$ and a heatmap $m$.

\paraheading{Baselines} We compare our model with other state-of-the-art models in the field of image-to-image translation, \emph{i.e.}, Pix2Pix~\cite{isola2016image}, BicycleGAN~\cite{zhu2017toward} and PG$^2$~\cite{ma2017pose}. Pix2Pix integrated an image conditioned GAN together with the $\ell_1$ loss for pixel-wise transformation. It can be seen as a base version of the proposed model without using the guidance of the facial mask, image fusion mechanism and diversity loss. BicycleGAN improved Pix2Pix by introducing conditional VAE~\cite{kingma2013auto} and latent regressor~\cite{donahue2016adversarial} for diversified image-to-image transformation, while in this work we achieve such an indeterministic transformation through the diversity loss. PG$^2$ proposed to explicitly introduce the body pose information into image-to-image generation. We implement all the baseline methods using their publicly released codes for a fair comparison.

Note that in Figure~\ref{fig-priorwork}, we have already shown the performance of geometric deformation based methods on our task. We conclude from the figure that although geometric deformation based methods can generate visually pleasing caricatures, the output is usually deterministic, lacking diversity in styles and exaggerations, while the GAN-based approaches can generate more diverse outputs. Hence in this section, we only compare our model with the state-of-the-art GAN-based models.

\begin{figure*}[!tp]
	\centering
	\includegraphics[width=1\linewidth]{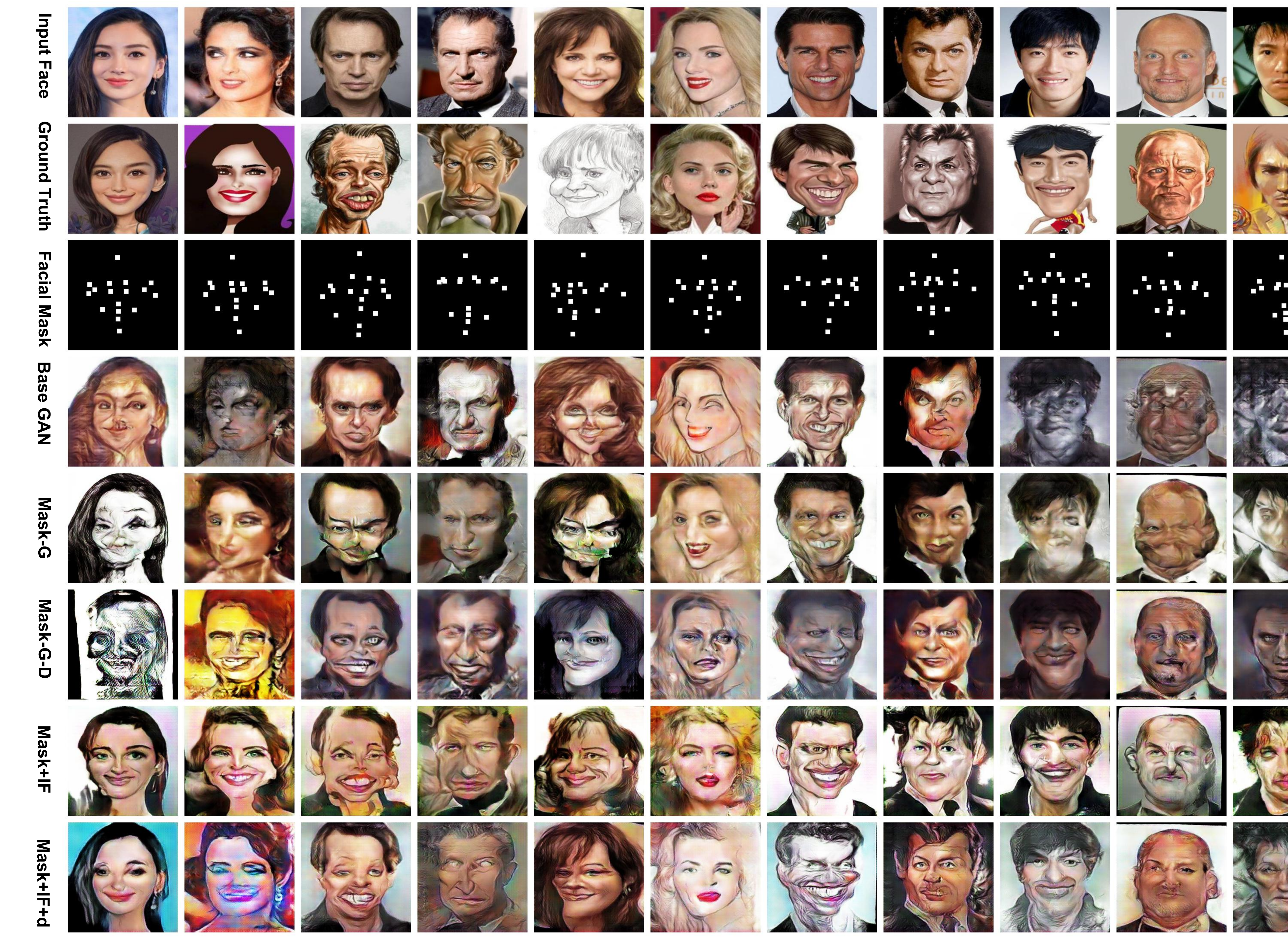}
	\caption{Qualitative comparison of different variants of the proposed model. From top to bottom: input, ground truth, facial mask, Base GAN, Mask-G, Mask-G-D and Mask+IF. We suggest the readers pay more attention to the facial parts, such as the eyes, mouth, \textit{etc}. Best viewed in color in screen with zoom-in.}
	\label{fig-ablation}
\end{figure*}

\begin{figure*}[!tp]
	\centering
	\includegraphics[width=1\linewidth]{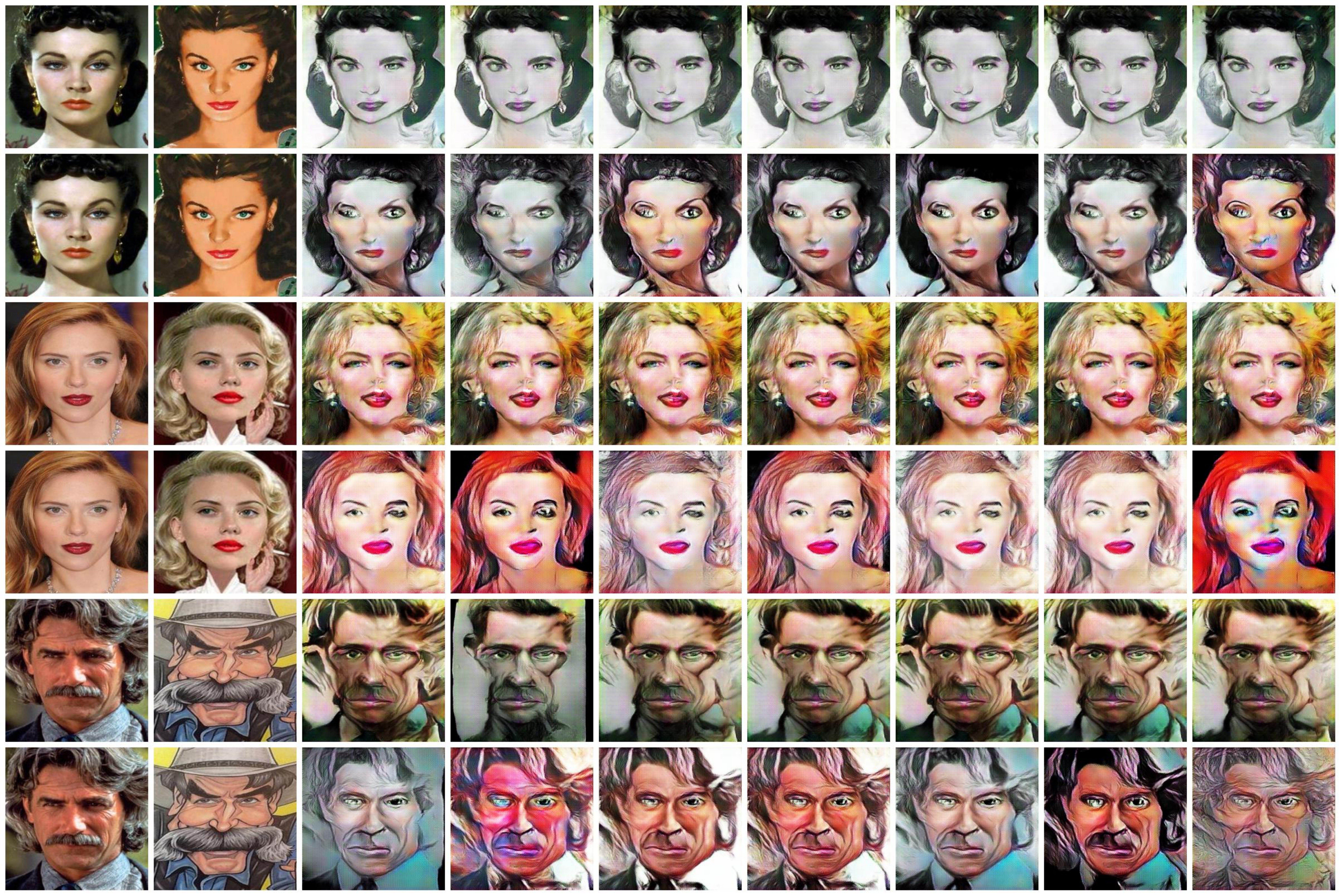}
	\caption{Interpolating on the noise $z$. The first and second columns are the input images and ground truth, respectively. The rest are the outputs with linearly interpolated noises. Odd rows: our model without diversity loss. Even rows: our model with diversity loss. Please pay special attention to the colors and textures of the generated caricatures.}
	\label{fig-div}
	\vspace{-1mm}
\end{figure*}

\paraheading{Implementation Details} We use a similar network as Pix2Pix~\cite{isola2016image}. The generator is a U-net like network which takes a random noise, a $256\times256$ image and a facial mask as input. The intermediate convolutional and deconvolutional layers are connected through skip-connections~\cite{he2016identity}. The discriminator is also composed of several convolutional layers. Each convolutional layer is followed by a Batch Normalization layer and a Leaky ReLU layer. Each deconvolutional layer is followed by a Batch Normalization layer \cite{ioffe2015batch} and a ReLU layer \cite{xu2015empirical}. We use Tanh as the activation function of the output layer of the generator and employ Sigmoid for the last layer of the discriminator. Adam is used as the optimizer to update the parameters of the entire model. In Adam, $\beta=0.5$ and the momentum is set to $0.9$. The learning rate is 0.0002 and is fixed during the training procedure.

\subsection{Ablation Study}
We first perform an ablation study to test the influence of each individual module of the proposed model. Specifically, we investigate the performance of the following models: Base GAN, Mask-G, Mask-G-D, Mask+IF and Mask+IF+diverse. Note that image fusion is called IF for short. Here, Base GAN is the model trained directly using cGAN and $\ell_1$ loss. It is essentially identical to the pix2pix model. Mask-G denotes the base GAN model with the facial mask as an additional input condition only to the generator $G$. Mask-G-D denotes the base GAN model with the facial mask as an input condition to both the generator $G$ and the discriminator $D$. Mask+IF is the facial mask conditioned model trained using the image fusion mechanism, \emph{i.e.}, $\mathcal{L}_{adv} + \mathcal{L}_{con}^{IF}$. As for the Mask+IF+diverse which is the full mode with $\mathcal{L}$, we will discuss it specifically in Section~\ref{Eva_diversity}. Note that for a fair comparison, all the models use the content loss to stabilize the training. 

The qualitative results of these models for the ablation study are shown in Figure~\ref{fig-ablation}. As can be seen, the Base GAN model without using any of the proposed designs gives perceptually the worst outputs. Especially, we notice that the outputs of the Base GAN model and the Mask-G model are aligned with the pose of the original input faces, while the results of the other models are well aligned with the given target caricature landmarks with reasonable exaggerations and correct viewpoints. This demonstrates that using a facial mask as the conditional information can help disentangle the exaggeration from other attributes of the faces and yield better exaggerated outputs. It also illustrates that the facial mask should be used for both generator $G$ and discriminator $D$. We can also observe that the Mask+IF model produce the best detailed outputs around the facial landmarks and hence the overall look of the generated caricatures are more realistic. This means the image fusion mechanism is indeed effective. It can help the generator to focus on generating images at key locations of the target subject. 

\subsection{Evaluation on Diversity Loss}
\label{Eva_diversity}
We further evaluate the effectiveness of the proposed diversity loss. To highlight the benefit of the diversity loss, we compare the visual difference between the outputs of the Mask+IF and Mask+IF+diverse models side by side under the same setting. Given a face image, we first randomly draw $7$ samples of $z$ from a Gaussian distribution. Then we feed the face image and the noise samples into the Mask+IF and Mask+IF+diverse models with the same facial mask. Figure~\ref{fig-div} shows the generated caricature images under different noises but with the same facial mask. Results demonstrate that the Mask+IF model produces deterministic outputs with negligible changes. Such an observation is consistent with previous work~\cite{isola2016image} about the noise ignorance problem. On the other hand, the outputs of the Mask+IF+diverse model are more diversified, showing that our diversity loss deals well with this problem. In the meanwhile, we can see vivid details from both results from the two models, indicating that our model is able to enhance the diversity without sacrificing the visual quality of the generated caricatures. 
\begin{figure*}[!tp]
	\centering	\includegraphics[width=0.75\linewidth]{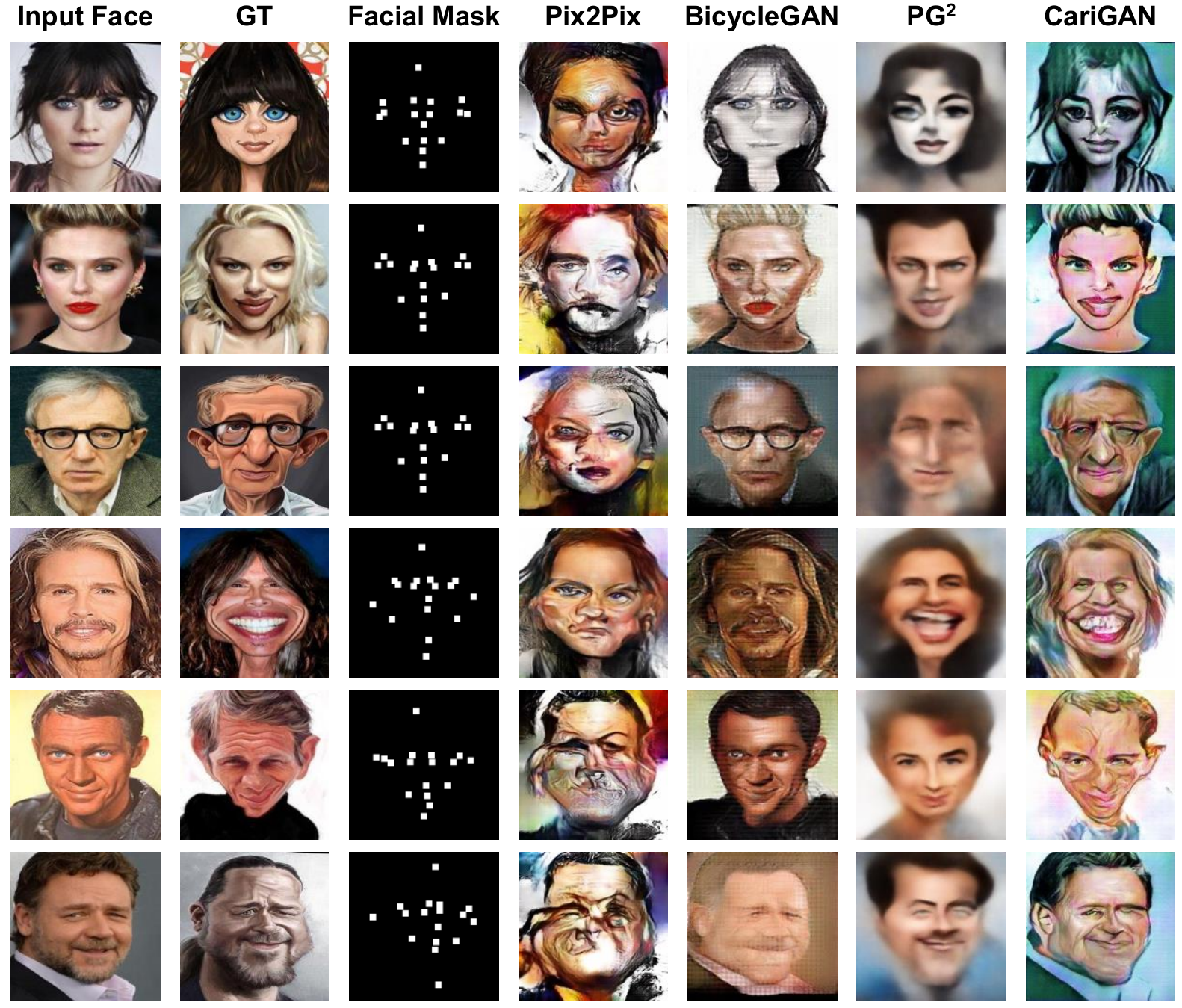}
	\caption{Qualitative comparison of our model against the baseline methods. From left to right: input, ground-truth, facial mask, results of Pix2Pix, BicycleGAN, PG$^2$, and our model. Please pay attention to the exaggerated facial regions, including eyes, mouth, nose, and so on.}
	\label{fig-base}
\end{figure*}

\subsection{Comparison with Baselines}
We also compare the proposed model with the state-of-the-art models. The qualitative comparison results are given in Figure~\ref{fig-base}. Please note that the ground-truth should only be used as {\it a reference} in the evaluation because {\it caricature generation is not a unique-solution problem with an exact pixel-level mapping}. In other words, there can be many plausible caricatures for a given face image. 

The figure reveals that due to the weakly-paired nature of our problem, models originally designed for pixel-wise image translation either cannot converge well, such as Pix2Pix, or generate somewhat identical images as the inputs, such as BicycleGAN. PG$^2$ produces images with better exaggerations with respect to the given caricature landmarks. However, as it heavily relies on the $\ell_1$ reconstruction loss, its outputs are blurry. In contrast, the outputs of our model are sharper. More importantly, they balance much better between the plausibility, identity and exaggeration, and therefore are visually much better than the results by the state-of-the-art methods.

In addition to the qualitative comparison, we also quantitatively measure the performance of our model against the state-of-the-art models in a user study. Since a face image may have multiple caricature counterparts, traditional evaluation metrics, such as SSIM and PSNR used for image-to-image translation models, are not applicable. Instead, we use human judgments for more perceptually reliable evaluation. We randomly pick $2$ face images for each person from the test set and obtain in total $104$ face images. For each face image, we generate a caricature image using the state-of-the-art models and our model. Then we ask $16$ participants to score the generated images. Each participant is assigned with 50 groups of images with each group containing the corresponding face image, ground truth caricature image and generated caricature image. The participants are required to score each generated image according to the following three aspects: (1) plausibility, whether the image is plausible enough; (2) identity preserving, whether the image has the same identity as the input face and the ground-truth caricature; (3) exaggeration, whether the generated image has the similar exaggeration (and we also ask them to check whether the viewpoint is correct) as the ground-truth caricature image. For each aspect, a caricature image receives a score between $1$ and $10$. We average the scores of all the participants.

The perceptual scores are given in Table~\ref{tab-perceptual-v}.  The Pix2Pix model receives the worst perceptual scores in all the three aspects. Because of the pixel-wise translation, BicycleGAN performs well in the plausibility and identity preserving aspects while showing poorly in the  exaggeration aspect. On the other hand, the PG$^2$ model addresses exaggeration well but has low scores in the plausibility and identity preserving aspects (due to the blurry outputs). Overall, our model performs well in all the three perceptual aspects and  produces the best all-around perceptual performance in terms of user rating scores.
\begin{table}[!tp] \small
	\centering
	\caption{Perceptual scores of different models, which are on a $10$-point scale. $10$ indicates that a perceptual aspect is best preserved and $1$ indicates a perceptual aspect is totally lost.}
	\begin{tabular}{@{}lcccc@{}}
	\toprule
	\textbf{Methods}	& \textbf{Plausibility} & \textbf{Identity}  &\textbf{Exaggeration}  & \textbf{Avg.} \\ \midrule
	\textbf{pix2pix}    & 1.92                  & 1.63               & 2.15                  & 1.90          \\
	\textbf{BicycleGAN} & 5.32                  & \textbf{6.91}      & 3.67                  & 5.30          \\
	\textbf{PG$^2$}     & 2.86                  & 2.85               & 5.08                  & 3.60           \\
	\textbf{Ours}       &\textbf{5.45}          & 5.34               & \textbf{6.18}         & \textbf{5.66}  \\ \bottomrule
	\end{tabular}
	\label{tab-perceptual-v}
\end{table}

\section{Conclusions}
We propose a CariGAN model based on conditional generative adversarial networks to address the four fundamental aspects of caricature generation task, \emph{i.e.}, Identity Preservation, Plausibility, Exaggeration and Diversity. Experiments demonstrate that using a facial mask as a condition of the cGAN model is crucial to the generation of appropriate exaggerations. It is also proved that the proposed image fusion mechanism can regularize our model to generate caricatures that are visually appealing in both global and local appearances. The diversity loss further encourages the model to produce diversified outputs given different random noise while preserving vivid exaggerations and accurate identity. Our model generates promising caricatures that handle all the four aspects of this task to a large degree, and clearly outperforms the state-of-the-art models in a user study. In the future work, we plan to further improve the performance in terms of those four aspects and extend the model to generate higher resolution caricatures.

{\small
\bibliographystyle{ieee}
\bibliography{references}
}

\end{document}